# Hair Segmentation and Skin Characterization on Macroscopic Dermatological Images for Laser Treatment Parameterization

**Alex Navarro Navarro**[1,2] · **Marc Arriaga Hidalgo**[1] · **Carlos Moreno Rubio**[1] · **Maxime Talarmain**[1] · **Noelia Vaquero Gallardo**[1,3]

Received: 4 May 2026 / Revised: 24 August 2026 / Accepted: 25 August 2026

**Abstract**
Current light-based devices, such as laser hair removal, intense pulsed light (IPL), vascular laser therapy, and dermatologic phototherapy, are widely used in clinical dermatology but rely heavily on subjective manual parameterization or uncalibrated point-based colorimeters. To overcome these limitations, this work presents a unified system for hair segmentation and skin characterization in macroscopic dermatological images. Unlike most existing literature that relies on dermoscopic datasets captured under highly controlled conditions, our approach is specifically designed for macroscopic images that replicate real clinical laser treatments. The proposed pipeline integrates a custom-trained variant of the YOLOv9 deep learning architecture for instance segmentation and classical image processing within a single workflow to extract objective skin and hair parameters such as phototype, hair density, color, and thickness. A color calibration stage using spectrophotometric measurements ensures accurate feature estimation regardless of illumination conditions. Experimental results show competitive segmentation metrics (mAP@50 $\approx$ 0.80) on a custom dataset featuring short, shaved hairs, typical of pre-treatment scenarios. These findings demonstrate the feasibility of the proposed pipeline in providing the objective, foundational skin and hair features required for future automated laser parameterization, paving the way for its extension to broader light-based medical therapies.

## Introduction

Laser hair removal is a widely used medical and aesthetic procedure that relies on the principle of selective photothermolysis of hair follicles. The effectiveness and safety of the treatment depend strongly on the correct adjustment of the laser parameters according to the patient's skin phototype, hair characteristics, and the presence of specific dermatological conditions [1]. In current clinical practice, this adjustment is typically performed manually by trained operators, based on visual inspection and experience. This manual process is time-consuming, subjective, and prone to variability between practitioners, which can affect both treatment outcomes and patient safety.

Accurate characterization of the treated skin area is essential: laser energy and pulse duration must be adapted to the patient's skin phototype and hair thickness to maximize efficacy while minimizing the risk of burns or hyperpigmentation. However, existing commercial devices rely heavily on human intervention for this task [2]. Only a few advanced systems integrate basic phototype estimation, often relying on simple point-based colorimeters that lack spatial context, or uncalibrated RGB cameras susceptible to lighting variations [3]. Automating this characterization process through computer vision would enable objective, consistent, and rapid parameter selection.

While deep learning has revolutionized dermatological image analysis, its application to aesthetic dermatology remains underexplored. Most existing computer vision approaches treat hair as an artifact to be removed for lesion diagnosis, rather than a clinical biomarker to be quantified.

✉ Noelia Vaquero Gallardo
noelia.vaquero@upc.edu

Alex Navarro Navarro
alejandro.navarro.navarro@estudiantat.upc.edu;
alexnavarro605@gmail.com

1 Intek Medical Technology, Barcelona, Spain

2 Facultat d'Informàtica de Barcelona (FIB), Universitat Politècnica de Catalunya (UPC), Barcelona, Spain

3 Department of Electronics Engineering, Eastern Barcelona School of Engineering (EEBE), Technical University of Catalonia – BarcelonaTech (UPC), Barcelona, Spain

Furthermore, they are typically designed for dermoscopic images under controlled lighting, failing to generalize to the macroscopic, uncalibrated images captured by laser handpieces in real clinical scenarios. In these settings, hairs are usually shaved—leaving only short, barely visible fragments—making segmentation notably more challenging.

To bridge this gap, we propose a unified pipeline where accurate instance segmentation is the first stage and directly supports subsequent skin and hair feature extraction. By treating hair as the primary signal and leveraging precise segmentation masks, we extract clinically relevant characteristics (such as hair density, thickness, color, and skin phototype) only on meaningful regions of the image. This localized approach prevents shadows or background noise from biasing global color measurements. The final output of the pipeline is a set of objective, quantitative features that can be used to inform laser treatment parameters.

Our contributions are threefold:

- We adapt and train a YOLOv9-based segmentation model for detecting and segmenting body hair in high-resolution macroscopic skin images captured through a laser device applicator, achieving competitive performance despite limited training data.
- We design a feature extraction pipeline that uses the segmentation output to precisely compute hair and skin characteristics (density, thickness, color, and phototype), integrating deep learning and classical image processing methods in a single workflow and addressing the limitations of manual visual estimation in clinical practice.
- We conduct a detailed evaluation on a custom dataset of macroscopic skin images, including a comparative study of color processing and calibration methods to ensure accurate and consistent feature estimation under real conditions.

## Related Work

### Hair Segmentation and Detection

The analysis of hair in dermatological images has traditionally been approached from the perspective of noise reduction. Early works relied on classical computer vision techniques such as morphological filtering and directional operators to detect and inpaint hair structures. Methods like DullRazor [4] or the matched filters proposed by Huang et al. [5] were specifically designed to erase hair from skin lesion images. Similarly, Yue et al. [6] proposed a classical hair counting method based on threshold segmentation, though limited to controlled microscopic scalp images.

More recent research has leveraged convolutional neural networks (CNNs) to improve hair detection. Architectures such as U-Net [7] and Mask R-CNN [8] have become standard for medical segmentation. For instance, Lama et al. [9] introduced ChimeraNet (an enhanced U-Net) for dermoscopic datasets, while Xiong et al. [10] proposed an attention-based model to detect sparse hair clusters. Despite these advances, the state of the art presents two major limitations for aesthetic applications. First, these models are overwhelmingly trained on dermoscopic datasets (e.g., ISIC) to assist melanoma diagnosis, struggling significantly when applied to macroscopic skin images where hair is shorter, less contrasted, and subject to uncontrolled lighting. Second, existing methods typically perform either basic object detection or semantic segmentation (binary hair vs. non-hair), but fail to unify both into instance segmentation. Our work diverges from this literature by utilizing a variant of the YOLOv9 architecture [11] to separate and measure each individual hair instance. Treating hairs as primary clinical targets rather than image artifacts is essential for accurate feature extraction and subsequent laser parameterization.

### Skin Analysis and Phototype Estimation

Objective skin phototype estimation is crucial to provide an objective proxy for subjective, self-reported clinical scales like the Fitzpatrick classification [12]. While the Fitzpatrick scale inherently incorporates biological reactivity to ultraviolet radiation, traditional computational approaches rely on the Individual Typology Angle (ITA) in the CIELAB color space [13, 14] to classify skin color continuously based purely on image-derived colorimetric indices, independent of any biological response, meaning they ultimately quantify distinct skin properties.

While recent deep learning studies have explored automatic skin tone classification [15–17], they frequently process images globally. Global analysis is highly susceptible to localized artifacts—such as hairs, shadows, or moles—which can severely distort the average color metrics. Moreover, most existing approaches assume well-calibrated acquisition settings. In real-world laser hair removal, images are captured using standard RGB cameras built into the handpiece, where automatic white-balancing and ambient light drastically alter the perceived skin tone. While some commercial solutions mitigate this using point-based colorimeters, these devices only measure a 1 $cm^2$ spot, ignoring the spatial distribution of melanin or local anomalies. Our work addresses these exact shortcomings by combining precise segmentation masks (to isolate clean skin) with a spectrophotometric calibration step, ensuring that the computed ITA values represent the true biological phototype and are invariant to the clinic's lighting conditions. Accurately extracting these features is clinically vital, as skin phototype and hair morphology directly dictate physical device irradiation settings—such as safe fluence thresholds and pulse durations governed by thermal relaxation times [18, 19]—effectively replacing the subjective, manual look-up tables common in current commercial systems.

## Materials and Methods

This section details the proposed pipeline for the objective characterization of skin and hair in macroscopic images. The methodology is structured into four main stages: first, the acquisition and annotation of the clinical dataset; second, the adaptation and training of the instance segmentation model (YOLOv9-C-DSEG) for hair detection; third, the spectrophotometric color calibration process to ensure illumination invariance; and finally, the mathematical extraction of clinical parameters, including hair density, thickness, and skin phototype.

### Dataset

A comprehensive dataset of high-resolution macroscopic dermatological images was compiled for hair segmentation and skin characterization. The images were acquired using a custom handpiece that integrates a Picamera2 sensor with LED illumination. This setup captures a 2 × 3 cm rectangular area of skin a few centimeters from the surface, simulating real-world conditions for laser hair removal treatments. They were acquired from volunteer participants during dedicated image acquisition sessions designed for research purposes. All volunteers signed informed consent authorizing the acquisition, anonymized storage, and scientific use of the images. The collected images were fully anonymized prior to analysis and contain no information that could directly identify the participants.

The dataset is composed of several subsets, each with distinct annotations and purposes. It includes a set of 384 initially labeled images. It was expanded with a new set of 1131 images, captured under improved illumination, of which 247 were annotated with refined criteria to enhance data quality, while the rest (884 images) were used for qualitative evaluation and testing. A separate set of 128 images was annotated by a domain expert to provide ground truth for detailed hair and skin characteristics (see Tables 1 and 2). Although the size of each subset was determined by volunteer availability, the total compiled data with hair annotations constitutes a sufficiently representative sample of the clinical variability (i.e., different skin phototypes and hair densities) required for robust model training and evaluation. However, the limited number of expert-annotated images produces imbalanced distributions across certain attributes, such as hair color and skin phototype.

The images depict various body regions, including arms, legs, armpits, and beards. All skin was shaved, clean, and dry to replicate a pre-treatment state. Images were acquired at a native resolution of 3840 × 2160 pixels (8-bit RGB) and stored in PNG format using lossless compression, ensuring that no image information was discarded prior to processing. After cropping to remove the handpiece's plastic borders, the final image resolution was 2918 × 1664 pixels, with every pixel corresponding to skin. The dataset captures a diverse population across different ages, sexes, skin tones, hair densities, and textures (although in variable proportions), which ensures the system's robustness to variations encountered in clinical practice.

With the assistance of a domain expert, an additional set of 128 labeled images was collected, annotated with relevant attributes such as body area, Fitzpatrick class, ITA color, hair color, hair thickness, and hair density. This dataset is intended to be used for testing and comparative evaluation. Note that for hair color, 8 images contained extremely fine hair where the expert could not reliably assign a color category; these images were excluded, resulting in 120 evaluated images for this specific feature.

Given the limited amount of annotated data, a 5-fold cross-validation was used for robust evaluation, with splits randomly partitioned by volunteer to ensure that all images from the same subject belonged to a single fold. No separate test split was used to preserve validation stability, as data scarcity directly limits model generalization.

### Annotation and Preprocessing

All images were meticulously annotated to ensure pixel-level accuracy of hair segmentation. A custom annotation tool integrating Segment Anything Model 2 (SAM2) [20] was employed to assist in generating initial masks, which were subsequently refined and manually verified by human annotators. No instance was accepted without human supervision, ensuring high annotation quality, which is critical for model performance.

A fixed ROI cropping was applied to all images to remove the visible plastic border of the acquisition handpiece, ensuring that the resulting images contain exclusively skin tissue. Additional preprocessing steps such as slicing, color correction, and data augmentation are described in their respective methodological sections.

### Hair Segmentation Model

The segmentation of body hair in macroscopic images poses unique challenges compared to conventional dermoscopic datasets. The hairs in our images are shaved, short, and low in contrast, often partially visible due to illumination and texture variability. Since the subsequent analysis requires quantifying individual hairs to estimate density, thickness, and color, the segmentation stage must preserve instance-level information rather than only pixel-wise foreground masks.

**Table 1** Summary of the datasets used in this study. *Other body regions in the expert-annotated dataset correspond to the arms (32), back (4), and chest (22)

| Dataset | Annotation | Images | Subjects | Sex | | Age (years) | | | | | Body region | | | |
|---|---|---|---|---|---|---|---|---|---|---|---|---|---|---|
| | | | | Female | Male | <30 | 31–40 | 41–50 | 51–60 | >61 | Cheeks | Legs | Axillae | Other |
| Initial dataset | Hair masks (initial quality) | 384 | 82 | Not collected | | | | | | | | | | |
| Expanded dataset | | 1131 | 208 | 111 | 94 | 50 | 43 | 57 | 41 | 15 | 340 | 407 | 384 | 0 |
| ↪ Refined | Hair masks (high quality) | 247 | 61 | 32 | 28 | 16 | 17 | 14 | 11 | 2 | 51 | 107 | 89 | 0 |
| ↪ Remaining | No annotations | 884 | 150 | 79 | 69 | 34 | 26 | 43 | 33 | 13 | 289 | 300 | 295 | 0 |
| Expert annotations | Skin phototype, hair color, thickness, and density | 128 | 18 | 5 | 13 | 7 | 7 | 4 | 0 | 0 | 18 | 38 | 14 | 58* |

**Table 2** Distribution of expert annotations for the 128-image evaluation dataset. Counts correspond to images. Hair color was evaluated on 120 images, as 8 images contained hair too fine to determine a color category

| Class | Count |
|---|---|
| Skin phototype | |
| I (Very light) | 28 |
| II (Light) | 36 |
| III (Intermediate) | 52 |
| IV (Tan) | 12 |
| V (Brown) | 0 |
| VI (Dark) | 0 |
| Hair color | |
| Blond | 8 |
| Brown | 66 |
| Black | 46 |
| Hair thickness | |
| Thin | 50 |
| Medium | 54 |
| Thick | 24 |
| Hair density | |
| Low | 40 |
| Medium | 64 |
| High | 24 |

Considering these requirements, we adopted the YOLOv9-C-DSEG architecture, a dual-head instance segmentation model that extends the standard YOLOv9-C architecture [11]. This variant, provided in the authors' official open-source repository,[1] incorporates an additional segmentation head that enhances the decoupling between object localization and mask generation, improving edge precision and mask stability without compromising inference speed. Compared with conventional semantic segmentation approaches, instance segmentation naturally provides independent masks for each detected hair, facilitating subsequent quantitative analysis. Furthermore, YOLO-based models offer an attractive trade-off between segmentation accuracy and computational efficiency, making them suitable for future real-time clinical applications. Within the YOLO family, preliminary experiments were conducted with GELAN-C-SEG, YOLOv9-C-SEG, and YOLOv9-C-DSEG under identical training conditions. Among these evaluated variants, YOLOv9-C-DSEG consistently produced the most accurate and stable segmentations and was therefore selected for the remainder of this work. A comparison of these models is presented in the "Hair Segmentation Performance" section.

The model was trained using the custom macroscopic skin dataset described previously, which includes a wide range of skin tones, hair densities, and anatomical regions. Training was performed with the dedicated YOLOv9-C-DSEG training pipeline, specifically designed for dual-head instance segmentation. Given the limited amount of annotated data available, data augmentation played a key role in effectively expanding the training set. Random geometric and photometric transformations—including rotations, flips, scaling, and controlled variations in brightness, contrast, and color were applied to increase the diversity of the samples and improve generalization. The segment copy-paste method was also used to generate more hair instances. Certain augmentation modes such as mosaic, mixup, perspective, and shear were explicitly disabled, as they degraded segmentation consistency on our macroscopic images, while translation and scaling were limited to small deformations to preserve structural realism. Empirical evaluation confirmed that applying controlled augmentation led to significantly better segmentation performance and more robust model behavior across heterogeneous skin conditions.

The model was trained on a workstation equipped with an MSI GeForce RTX 4090 SUPRIM X GPU (24GB GDDR6X). Training was conducted for 300 epochs with a batch size of 8 and 8 parallel workers, using Stochastic Gradient Descent (SGD) as the optimizer and an initial learning rate scheduler without cosine decay (cos_lr = false). An early stopping patience of 20 epochs was used to prevent overfitting and ensure stable convergence, using a fitness function weighting 0.1 mAP@50 and 0.9 mAP@50-95 (segmentation), as is typical in the YOLO training pipeline.

The final model was trained exclusively on the subset of 247 images with high-quality, refined annotations, as preliminary experiments showed that including the larger set of initial, lower-quality annotations degraded performance, highlighting the critical importance of annotation quality over dataset size for this task. To ensure robust evaluation, a 5-fold cross-validation strategy was employed. The 247 images, corresponding to 61 unique volunteers, were randomly partitioned by subject into five folds to prevent data leakage: four folds with images from 12 volunteers and one with 13, resulting in image counts of 52, 49, 48, 53, and 45 per fold. For both training and inference, high-resolution images are divided into 18 overlapping $640 \times 640$ tiles with a 20% overlap ratio to preserve detail and fit into GPU memory, with pixel values normalized to the [0, 1] range. This tiling process resulted in training sets of 3510, 3564, 3582, 3492, and 3636 tiles, and corresponding validation sets of 936, 882, 864, 954, and 810 tiles for each of the five folds. During inference, each tile is processed independently, and the resulting detections and masks are merged back into the full image. Overlapping detections are handled through non-maximum suppression (NMS) with a confidence threshold of 0.20 and an IoU threshold of 0.45, followed by a containment filter. This filter removes redundant detections by considering a box contained if 65% of its area is enclosed within another, resulting in a single, coherent segmentation map. A comprehensive list of all training hyperparameters is provided in Table 3.

[1] https://github.com/WongKinYiu/yolov9

## Color Calibration

Accurate color representation is essential for all subsequent analyses related to skin and hair characterization. To ensure that the pixel values in the captured images correspond as closely as possible to the true observed colors, a color calibration procedure was carried out using a spectrophotometer in combination with a Pantone SkinTone™color palette, which provides a representative sampling of human skin tones across a wide chromatic range.

Directly measuring the spectral response of human skin with a spectrophotometer is not reliable due to the complex optical properties of the skin (specifically subsurface light scattering and absorption) [3], which produce inaccurate reflectance readings. To overcome this limitation, calibration was performed using the Pantone color samples, which provide stable and reproducible surface reflectance. For each of the 30 selected color patches, an image was captured using the same optical setup and illumination conditions as those used for the skin dataset. Additionally, an ROI cropping has been applied to retain only the center of the image, which exhibits the most uniform illumination. The corresponding ground-truth color for each patch was measured in the CIELAB space with a calibrated spectrophotometer, producing a paired dataset of camera-measured and reference color values.

The selected color correction algorithm is based on the "Color Transfer between Images" [21] approach. Although originally proposed for color style transfer, this method has been widely adopted for practical color correction. It computes a statistical transformation between a source color distribution (the ground-truth measurements) and a target distribution (the camera-captured values) by aligning their means and standard deviations in the LAB color space. Applying this transformation to the target images effectively maps their chromatic properties to match the reference standard.

To assess the performance of the calibration, the mean Euclidean distance in CIELAB space ($\Delta E$) between the measured and reference colors was computed before and after correction. The mean color error decreased from 26.20 to 4.58 after applying the Reinhard correction, indicating a substantial improvement in chromatic accuracy.

For comparison, additional color correction methods were evaluated. A linear $3\times3$ correction matrix in the XYZ space [22, 23] reduced the mean error from 26.20 to 6.62, while a second-degree polynomial regression model achieved an apparent training error of 2.64 but exhibited overfitting, with a validation error increasing to 9.25 on unseen colors. These results confirm that the Reinhard-based correction provides the best trade-off between accuracy, generalization, and computational efficiency for this imaging setup.

This calibrated color mapping is applied to all captured skin images after the segmentation stage but prior to any feature extraction. Consequently, it does not affect segmentation performance and is used exclusively to ensure that subsequent estimations of skin tone, phototype, and hair color are based on perceptually consistent and physically meaningful color data.

## Skin Color

The analysis of skin color was designed to ensure both generalization and robustness across individuals and imaging conditions. For this purpose, we adopted the Individual

**Table 3** Key training hyperparameters for the YOLOv9-C-DSEG model. Parameter names correspond to the standard YOLOv9 configuration

| Param. | Value | Param. | Value | Param. | Value |
|---|---|---|---|---|---|
| Initial LR (lr0) | 0.01 | Object loss (obj) | 0.7 | Translation (translate) | 0.1 |
| Final LR (lrf) | 0.01 | Object PW (obj_pw) | 1.0 | Scaling (scale) | 0.1 |
| Momentum | 0.937 | DFL loss (dfl) | 1.5 | Shear | 0.0 |
| Weight decay | 0.0005 | IoU thresh. (iou_t) | 0.20 | Perspective | 0.0 |
| Warmup epochs | 3.0 | Anchor thresh. (anchor_t) | 5.0 | Flip (up-down) | 0.5 |
| Warmup momentum | 0.8 | FL gamma (fl_gamma) | 0.0 | Flip (left-right) | 0.5 |
| Warmup bias LR | 0.1 | HSV-Hue (hsv_h) | 0.015 | Mosaic | 0.0 |
| Box loss (box) | 7.5 | HSV-Sat (hsv_s) | 0.7 | Mixup | 0.0 |
| Class loss (cls) | 0.5 | HSV-Val (hsv_v) | 0.4 | Copy-paste | 0.3 |
| Class PW (cls_pw) | 1.0 | Rotation (degrees) | 0.0 | | |

Typology Angle, a well-established colorimetric descriptor that provides an objective and standardized measure of skin tone. The ITA is computed from the lightness ($L^*$) and blue–yellow chromatic component ($b^*$) of the CIELAB color space according to the following equation:

$$ITA = \arctan\left(\frac{L^* - 50}{b^*}\right) \times \frac{180}{\pi} \quad (1)$$

This value, expressed in degrees, can be discretized into six phototype classes commonly used in dermatological studies (very light, light, intermediate, tan, brown, and dark) using standardized thresholds (e.g., 55°, 41°, 28°, 10°, and –30°) [13]. Before computing the ITA, pixel values are filtered to improve stability: pixels with $L^* = 0$ are discarded, and only those within one standard deviation from the mean of both $L^*$ and $b^*$ are retained to minimize the influence of noise and local artifacts.

When visualizing our dataset in the $L^*$–$b^*$ space and overlaying the discrete ITA thresholds (Fig. 1), the distribution of skin tones follows the characteristic curved "tear-drop" shape originally described in Chardon's work, confirming the expected chromatic structure of human skin under controlled illumination (Fig. 2).

However, despite the apparent simplicity of the ITA formula, several practical factors complicate its estimation in macroscopic imaging. At the magnification level used in our system, subtle surface irregularities (such as pores, fine wrinkles, and micro-shadows) introduce perceptible local color variations even within seemingly uniform regions. Moreover, an illumination vignette effect can produce brightness gradients, typically making the central region appear lighter while peripheral areas darken progressively. As a result, the distribution of ITA values in a single image can span multiple phototype categories, sometimes up to four or five adjacent classes.

To mitigate these issues, a Region of Interest (ROI) is defined at the center of the image, corresponding to the same calibrated area used during the color correction procedure. This ensures that only pixels within the most color-accurate region (where the camera's chromatic response is well characterized) are considered for phototype estimation. Additionally, the hair segmentation mask is applied to exclude hair-covered pixels, further reducing chromatic contamination.

Together, these steps—color calibration, central ROI selection, and segmentation-based filtering—ensure that the resulting ITA value and its corresponding phototype class reliably represent the true skin color in the imaged area, enabling consistent and reproducible estimation across individuals and acquisition sessions.

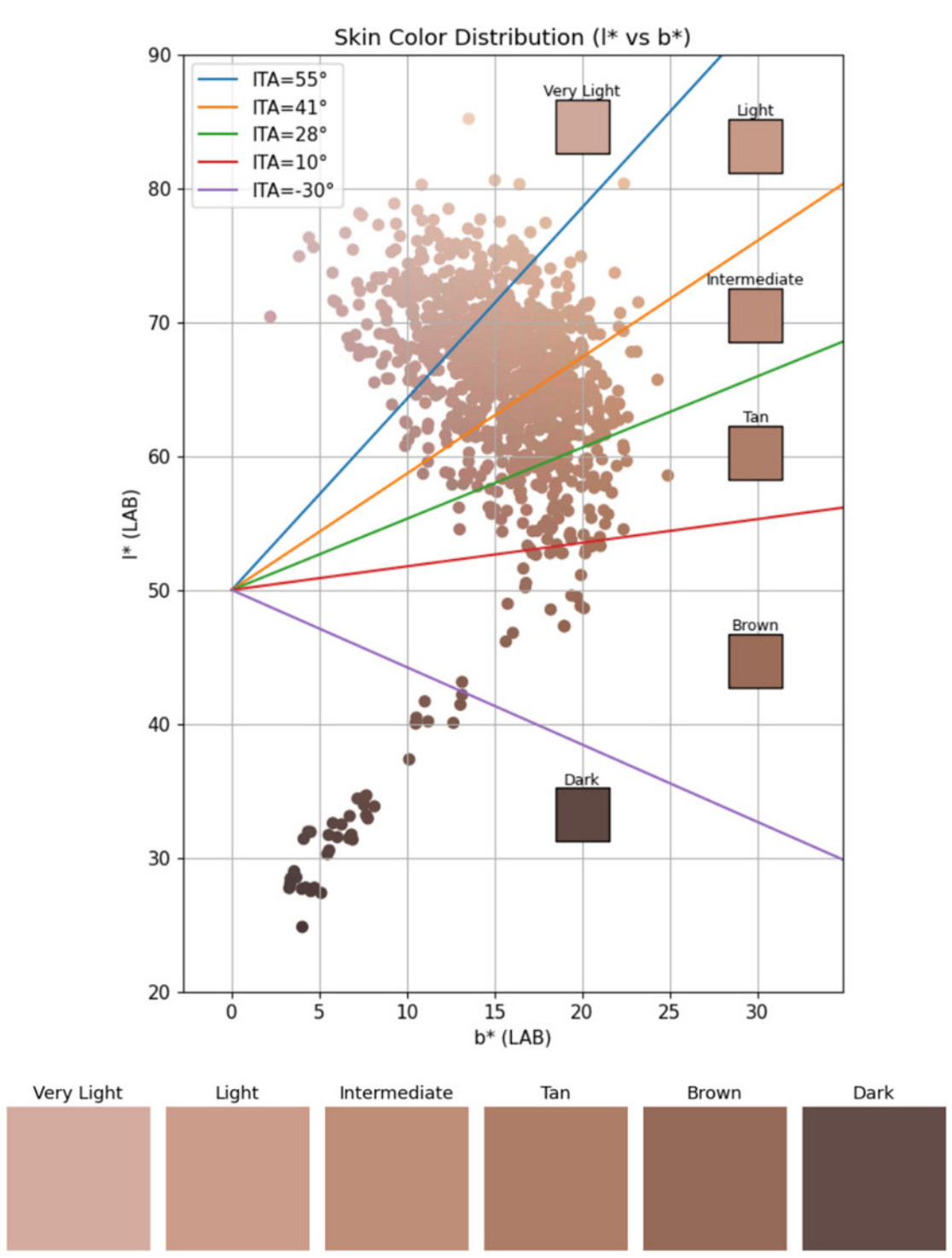

**Fig. 1** Skin color distribution of our dataset in the $L^*$–$b^*$ space with ITA discretization boundaries overlaid (top) and mean color per ITA class (bottom). Each point represents the average skin color of a single image from the expanded dataset (1131 images)

## Hair Color

Estimating hair color from macroscopic skin images presents notable challenges. Unlike skin tone, for which standardized quantitative colorimetric scales exist (e.g., ITA in the CIELAB space), there is no universally accepted metric for hair color. Additionally, due to the fine and partially translucent nature of body hair, as well as the optical blending between hair and skin at submillimeter scales, the extracted color values are often influenced by surrounding skin reflectance.

The color of the hair is computed directly from the segmentation masks produced by the model. Each mask is applied to the original RGB image to isolate only the pixels corresponding to detected hairs. However, these masked regions typically include mixed pixels at the hair–skin boundary, where partial transparency, focus blur, and color bleeding cause contamination by skin tones. As a result, directly averaging all hair pixels yields biased color estimates that are systematically lighter than the true hair color.

To mitigate this problem, two different strategies were evaluated:

- Erosion-based refinement: A morphological erosion is applied to each hair mask to remove boundary pixels, retaining only the central region of the hair where color contamination is minimal. Although simple and computationally efficient, this method may excessively

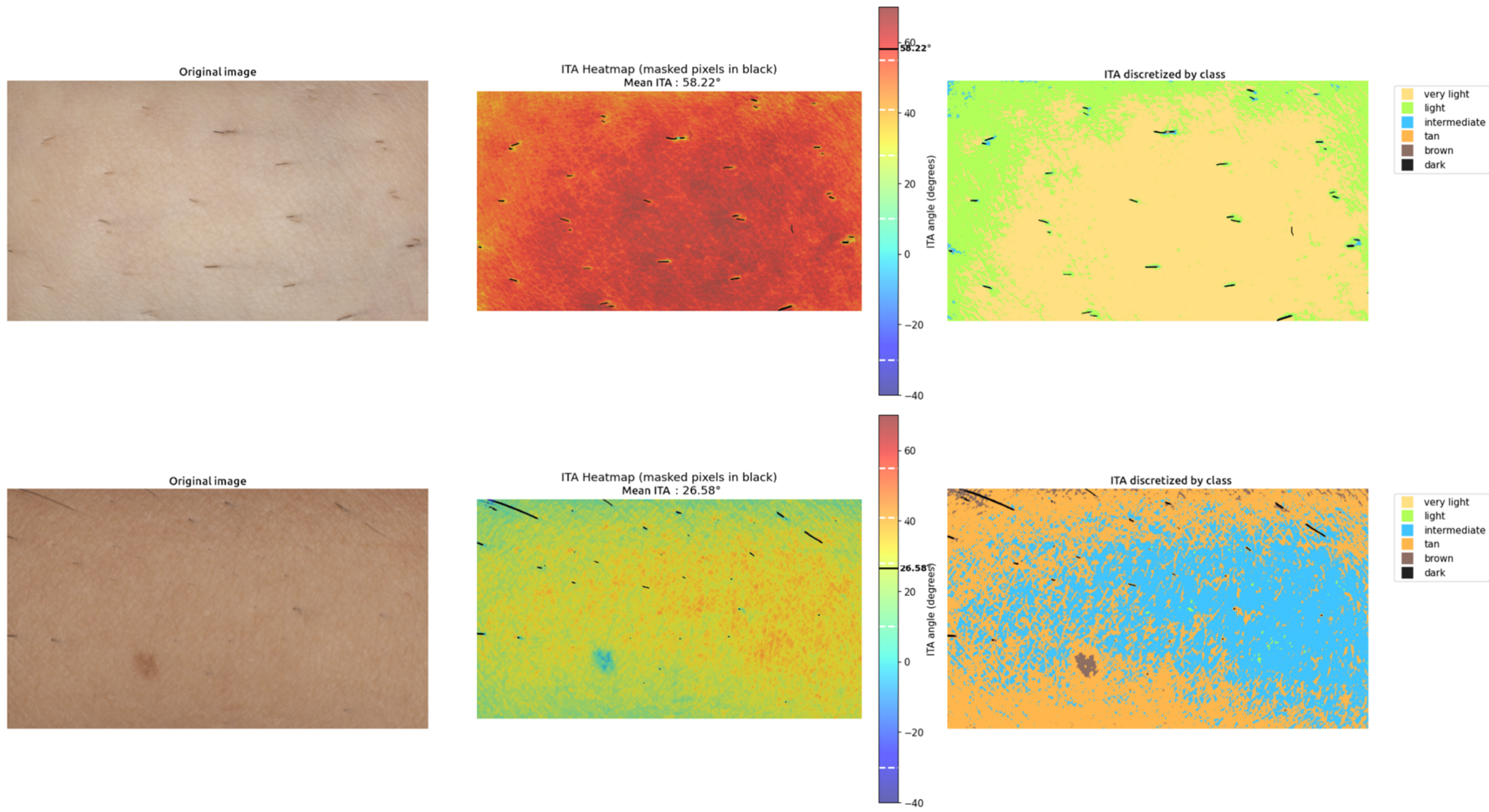

**Fig. 2** Visualization of the ITA values and class distribution in different images. Images correspond to: (left) input image, (center) ITA value, and (right) discretized ITA class

shrink thin hair masks, potentially discarding a large proportion of valid pixels, particularly for very fine hairs.

- Clustering-based color separation: A more robust but computationally heavier approach involves performing color clustering on the combined set of hair and skin pixels (e.g., using $k$-means with $k = 6$) in the RGB color space. The cluster with the lowest proportion of skin pixels is identified as corresponding to hair, and the centroid of that cluster is taken as the representative hair color. To maintain computational feasibility, a random sample of 20000 pixels from each region is used in the clustering step.

As illustrated in qualitative comparisons (Fig. 3), both methods effectively darken the estimated hair color relative to the uncorrected mean, aligning more closely with visual perception. The clustering-based approach generally yields better results for medium and thick hairs, while very thin hairs remain partially affected by skin color interference due to the limited number of pure hair pixels within their masks.

Once the hair color noise is filtered using either of these methods, an unsupervised classification is applied to categorize the hair tone into the three predominant color groups commonly used in the field (blonde, brown, and black). This is achieved by performing a second $k$-means clustering with $k = 3$ in the CIELAB color space on the resulting mean hair color values across all the images of the whole dataset. To ensure robust convergence, both this and the previous clustering stage employ the k-means++ algorithm for centroid initialization. The centroids of these clusters are stored as reference tones and can later be used for the automatic classification.

This two-stage strategy provides a reliable estimation of both individual and categorical hair color. By first removing skin-induced color artifacts and then clustering hair tones in a perceptually consistent manner, the system achieves robust color characterization suitable for integration into the overall laser treatment parameterization workflow.

## Hair Count and Density

The estimation of hair count and follicular density is straightforward thanks to the instance segmentation approach described in the “Hair Segmentation Model” section. Each detected instance corresponds to a single hair, allowing direct counting without the need for additional component analysis or post-processing. Even in cases where hairs are partially overlapping or crossing—although uncommon in the captured dataset—the model successfully distinguishes them as separate instances. Furthermore, the reconstruction procedure used to merge predictions from overlapping image tiles ensures that no duplicated detections are present, resulting in accurate and consistent hair counts across the full-resolution image.

To compute follicular density, typically expressed as the number of hairs per square centimeter, a precise pixel-to-millimeter conversion factor was experimentally determined. A calibration pattern with known physical dimensions was photographed under the same optical setup and acquisition conditions as the skin images. By comparing the number of pixels corresponding to the calibrated area, the system's spatial resolution in millimeters per pixel was obtained (0.00732318 mm/pixel). This conversion enables the transformation of the detected hair count into a physically meaningful density value (hairs/mm$^2$ or hairs/cm$^2$), directly comparable to standard dermatological measures (Fig. 4).

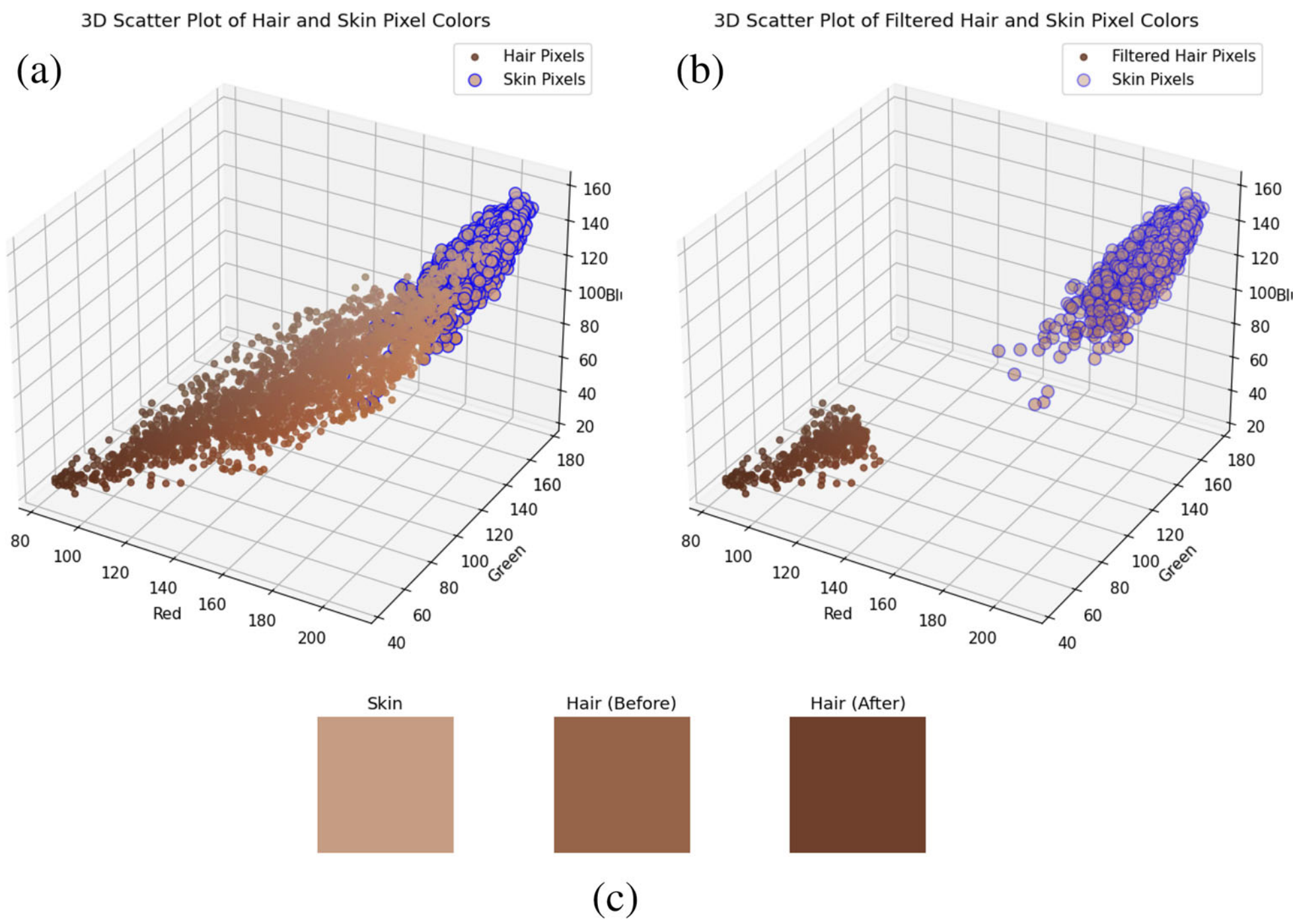

**Fig. 3** Visualization of skin color contamination in hair pixel masks and the effect of the proposed clustering-based filtration. **a**) Overlaid hair and skin pixels in RGB space indicating color contamination. **b**) The dominant cluster, corresponding to filtered hair pixels after k-means separation. **c**) Comparison of estimated hair color before and after clustering-based filtration, demonstrating reduced skin-induced bias

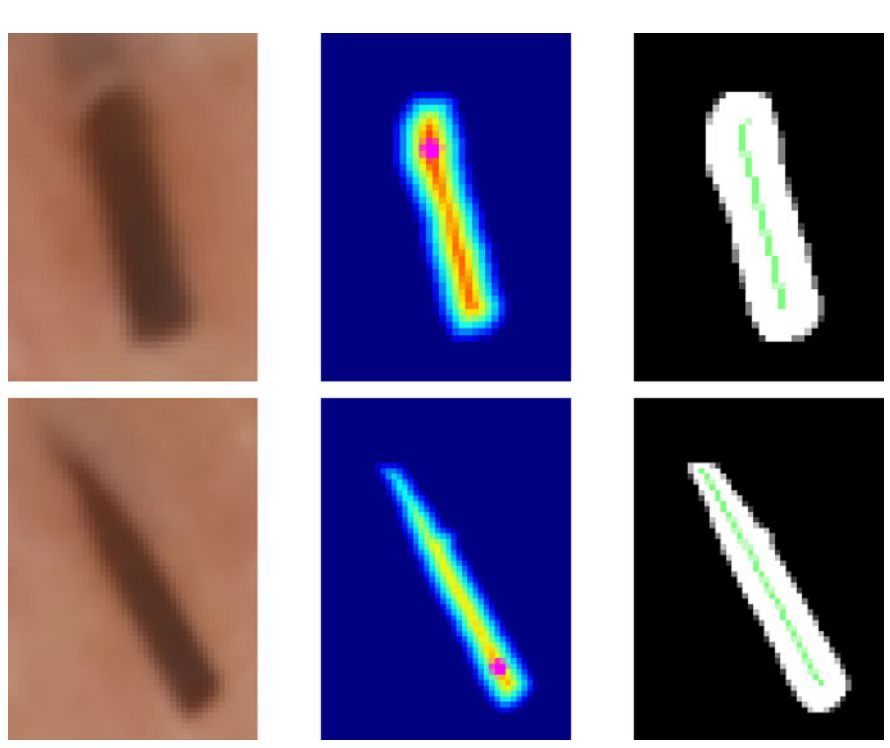

**Fig. 4** Visualization of the transformations used for hair thickness estimation. Original image (first column), distance transform of the mask (second column), and skeleton of the mask (third column)

## Hair Thickness

The estimation of hair thickness is performed using classical image processing techniques applied to the binary masks produced by the segmentation model. For each detected hair instance, the corresponding mask is processed using the distance transform, which computes, for every pixel within the mask, its distance to the nearest boundary pixel. This transformation provides a spatial representation of the local hair radius and allows two complementary thickness measures to be derived:

- Maximum hair thickness: The maximum value of the distance transform within the hair mask represents the radius at the thickest point of the hair. Multiplying this value by two yields the maximum hair thickness.
- Mean hair thickness: By extracting the skeleton of the hair mask—the central axis representing the medial line of the structure—and averaging the distance transform values along this skeleton, an accurate approximation of the mean hair thickness is obtained.

Both measurements are initially expressed in pixels. To convert them into physically meaningful units, the same pixel-to-millimeter conversion factor described in the previous section is applied. This calibration ensures that the estimated hair diameters are reported in millimeters and can be directly compared with dermatological and trichological standards. The combination of distance transform and skeleton-based analysis provides a robust and computationally efficient method to quantify the fine structural characteristics of hair in macroscopic skin images.

## Experiments and Results

### Hair Segmentation Performance

The performance of the three candidate models (YOLOv9-C-DSEG, YOLOv9-C-SEG, and GELAN-C-SEG) was evaluated using a 5-fold cross-validation strategy, with the averaged results presented in Table 4. Performance metrics followed the standard COCO evaluation protocol [24], reporting mean average precision (mAP) at Intersection-over-Union (IoU) thresholds of 0.5 and 0.5:0.95. The results show a highly competitive and remarkably similar performance among the models, with differences of less than 0.01 in any given metric. All models achieve an mAP@50 of approximately 0.80, indicating robust localization, while the mAP@50-95 scores (around 0.47 for detection and 0.45 for segmentation) reflect the inherent challenge of achieving pixel-perfect boundaries on fine structures.

To provide a more granular assessment of segmentation quality, particularly concerning mask boundary precision, a detailed analysis using region- and boundary-based metrics was conducted, with results summarized in Table 5. This evaluation addresses the concern that while mAP scores are robust, a more detailed look at mask fidelity is needed, especially since boundary accuracy directly influences subsequent hair thickness estimation. The results reveal a consistent pattern where micro-averaged metrics are notably higher than their macro-averaged counterparts (e.g., Micro Dice of 0.8332 vs. Macro Dice of 0.7710). This discrepancy is explained by the averaging method and the nature of our data: macro-averaging gives equal weight to each image, so images with challenging conditions, such as those with very few or extremely fine hairs, can disproportionately lower the average score if they contain errors. In contrast, micro-averaging aggregates performance across all pixels globally, better reflecting the model's overall strong performance on the majority of hair instances.

The main factors contributing to lower scores in specific images are twofold. First, the occasional misidentification of small, hair-like artifacts on the skin, such as clothing fibers or debris, which are incorrectly segmented as hair. Second, the inherent difficulty in precisely delineating the boundaries of very fine hairs (e.g., 4–5 pixels wide). For these instances, a minimal one-pixel error caused by image blur can lead to a high relative error, whereas the same absolute error has a negligible effect on thicker hairs. Despite these challenges, the high micro-averaged scores (e.g., micro F1-score of 0.8332) and boundary F1-score (0.8153) confirm that the model produces accurate masks at a global level. The impact of these minor boundary inaccuracies on the final hair thickness estimation is quantitatively analyzed in a subsequent section and shown to be minimal.

Based on this detailed analysis, YOLOv9-C-DSEG was selected as the final model. While the standard COCO metrics (Table 4) showed a nearly identical performance across all candidates, with differences smaller than 0.01, the more granular evaluation in Table 5 reveals a clearer distinction. YOLOv9-C-DSEG consistently outperforms the other models in most region-based (Dice, IoU) and boundary-based metrics, especially in the macro-averaged scores, which are more sensitive to challenging images. Since the accuracy of these masks is critical for the subsequent feature extraction stages, the superior performance of YOLOv9-C-DSEG in these specific metrics justifies its selection for the final pipeline.

Figure 5 presents representative examples of the model's inference outputs across different skin phototypes, hair densities, and anatomical regions, illustrating both successful segmentations and typical system limitations. The visualization displays the original macroscopic input images, the predicted bounding boxes and instance contours, and the final extracted binary masks. Discrepancies or false detections in these visual examples are primarily located in areas containing specular reflection glare or low-contrast hair stubs, with a notable decrease in segmentation quality visible in the fourth image (very dark skin tone) compared to the first three examples (lighter skin tones).

**Table 4** Comparative quantitative performance metrics for different segmentation models. Results are reported as mean ± standard deviation across the validation sets of a 5-fold cross-validation. Best results are highlighted in bold

| Task | Model | Precision | Recall | mAP@50 | mAP@50–95 |
|---|---|---|---|---|---|
| Detection (box) | YOLOv9-C-DSEG | **0.7942 ± 0.0185** | 0.7411 ± 0.0220 | 0.7943 ± 0.0168 | 0.4714 ± 0.0137 |
| | YOLOv9-C-SEG | 0.7871 ± 0.0194 | 0.7443 ± 0.0214 | 0.7964 ± 0.0241 | 0.4750 ± 0.0177 |
| | GELAN-C-SEG | 0.7877 ± 0.0159 | **0.7455 ± 0.0158** | **0.7992 ± 0.0175** | **0.4811 ± 0.0164** |
| Segmentation (mask) | YOLOv9-C-DSEG | 0.7875 ± 0.0170 | 0.7242 ± 0.0261 | 0.7965 ± 0.0128 | 0.4430 ± 0.0072 |
| | YOLOv9-C-SEG | 0.7786 ± 0.0203 | **0.7293 ± 0.0227** | 0.7970 ± 0.0168 | **0.4517 ± 0.0058** |
| | GELAN-C-SEG | **0.7885 ± 0.0097** | 0.7195 ± 0.0247 | **0.7984 ± 0.0118** | 0.4511 ± 0.0070 |

**Table 5** Comparative detailed evaluation metrics for hair segmentation masks across different models on the validation set. Macro Mean averages the metric per image, while Micro Mean computes it globally. Best results are highlighted in bold

| Metric | Type | YOLOv9-C-DSEG | YOLOv9-C-SEG | GELAN-C-SEG |
|---|---|---|---|---|
| Dice score | Macro | **0.7710** | 0.7405 | 0.7076 |
| | Micro | **0.8332** | 0.8297 | 0.8071 |
| IoU | Macro | **0.6467** | 0.6179 | 0.5832 |
| | Micro | **0.7140** | 0.7090 | 0.6766 |
| Pixel precision | Macro | **0.7868** | 0.7415 | 0.6901 |
| | Micro | **0.8128** | 0.8090 | 0.7678 |
| Pixel recall | Macro | 0.7816 | 0.7857 | **0.7983** |
| | Micro | **0.8546** | 0.8516 | 0.8506 |
| Pixel F1-score | Macro | **0.7710** | 0.7405 | 0.7076 |
| | Micro | **0.8332** | 0.8297 | 0.8071 |
| Boundary precision | Macro | **0.8013** | 0.7641 | 0.7188 |
| | Micro | 0.8054 | **0.8073** | 0.7578 |
| Boundary recall | Macro | 0.7806 | 0.7619 | **0.8060** |
| | Micro | **0.8253** | 0.8237 | 0.8189 |
| Boundary F1-score | Macro | **0.7758** | 0.7505 | 0.7251 |
| | Micro | 0.8153 | **0.8154** | 0.7872 |

Inference was executed on an NVIDIA GeForce RTX 4060 Laptop GPU (CUDA 12.1) without batch processing, meaning that further optimization or deployment on higher-end hardware could achieve faster runtimes. Under the current configuration, the average processing time was 0.0453 s per tile and 0.8156 s per full reconstructed image, including slicing and post-processing steps.

Despite overall robust segmentation across diverse skin tones, performance notably decreases for individuals with very dark skin. In these cases, the lack of visible contrast or gradient between hair and surrounding skin prevents reliable boundary detection, as both appear chromatically similar. This limitation highlights the need for alternative modalities—such as multispectral imaging or polarization cues—to improve segmentation robustness in darker phototypes.

## Error Propagation to Feature Estimation

To quantify the impact of segmentation inaccuracies on the final clinical parameters, we analyzed the error propagation from the segmentation masks to the derived hair thickness and density estimations. This was done by comparing the features extracted from the model's predicted masks against the features extracted from the ground truth masks. Although the mask evaluation metrics in the previous section indicated some room for improvement, the results in Table 6 demonstrate that these minor boundary imperfections have a negligible impact on the derived features.

For mean hair thickness, the model achieves a remarkably low macro-averaged absolute error of just 4.0 μm, which corresponds to a sub-pixel deviation. This is further supported by the low relative error (macro: 7.80%, micro: 0.02%), confirming that the segmentation boundaries are sufficiently precise for reliable thickness estimation.

For hair density, the discrepancy between macro- and micro-averaged relative errors (33.58% vs. 4.11%) is again observed, attributable to the disproportionate impact of errors in images with low hair counts, as discussed previously. The micro-averaged error, which reflects global performance, is a more representative indicator and, at 4.11%, demonstrates strong overall accuracy. Since this evaluation compares predictions against the high-quality, refined ground truth masks, these low error values validate that the proposed pipeline yields reliable quantitative features.

## Feature Extraction Evaluation

To quantitatively validate the clinical relevance of the extracted features, the performance of the feature extraction modules was evaluated against the expert-annotated subset of 128 test images. This ground truth dataset provides

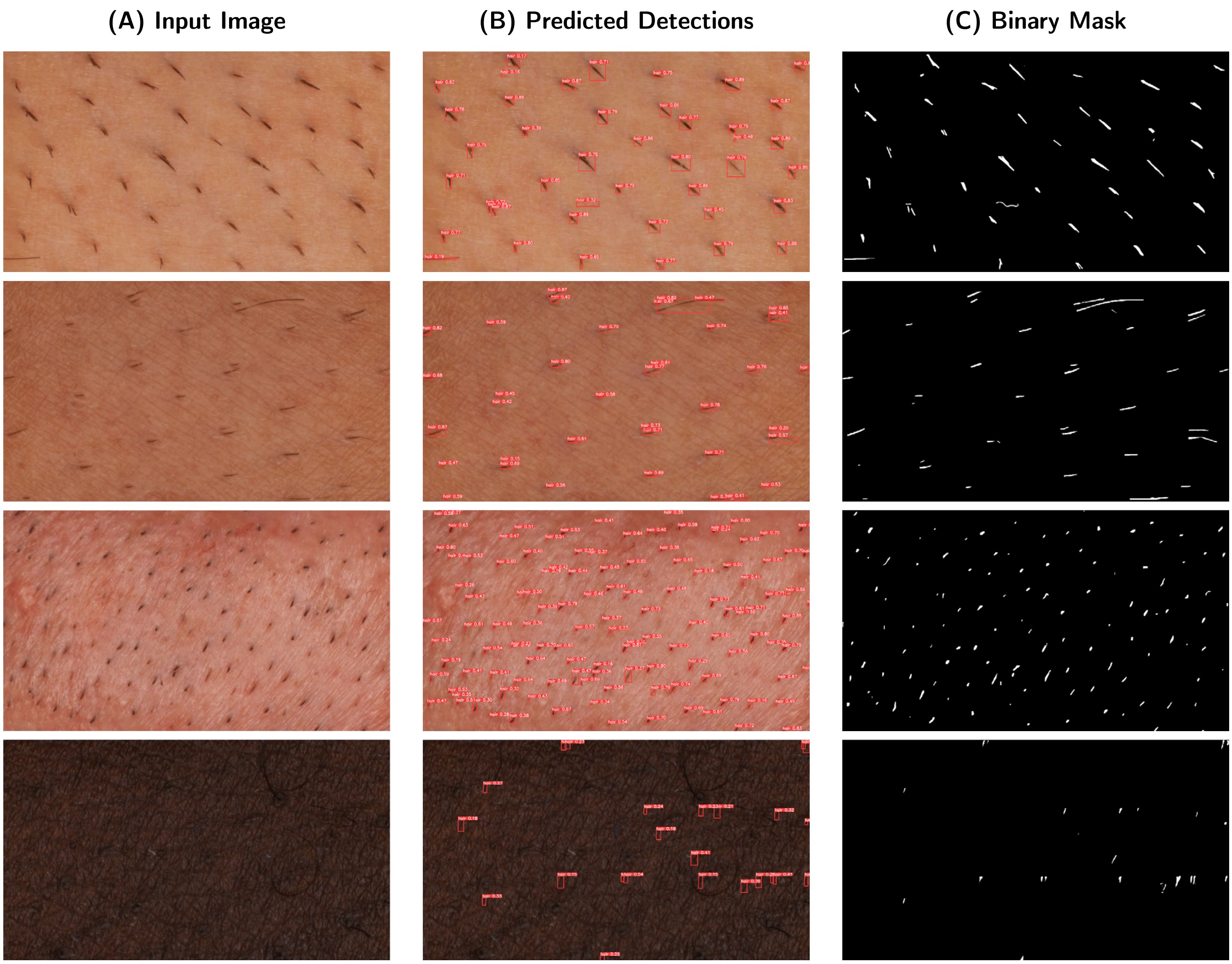

**Fig. 5** Segmentation model inference results across four distinct test samples (rows). **A**) Original macroscopic input images. **B**) Predicted hair detections. **C**) Extracted binary mask visualizations

categorical labels for skin phototype, hair color, density, and thickness.

A key challenge in this evaluation is comparing the system's continuous numerical outputs for hair thickness and density against the expert's categorical annotations (e.g., "thin," "medium," "thick"). Since no standardized thresholds exist for these features and their definitions vary significantly by application, we discretized the model's predictions using thresholds defined by the expert annotator for this specific dataset (thickness: 0.05 and 0.07 mm; density: 0.10 and 0.30

**Table 6** Evaluation of hair thickness and density estimation against ground truth masks for the YOLOv9-C-DSEG model on the 247-image refined annotation dataset. Errors are reported as both Macro (per-image average) and Micro (global average) means

| Parameter | Metric | Macro mean | Micro mean |
|---|---|---|---|
| Mean hair thickness | Error (px) | −0.0170 | 0.0017 |
| | Abs. error (px) | 0.5480 | 0.0017 |
| | Error (mm) | −0.0001 | 0.00001 |
| | Abs. error (mm) | 0.0040 | 0.00001 |
| | Rel. error (%) | 7.80 | 0.02 |
| Hair density | Error (hairs/mm$^2$) | 0.0052 | 0.0053 |
| | Abs. error (hairs/mm$^2$) | 0.0183 | 0.0053 |
| | Rel. error (%) | 33.58 | 4.11 |

**Table 7** Quantitative performance metrics for feature extraction modules evaluated against expert annotations on the test set ($N = 128$, except for Hair Color where $N = 120$). Metrics are averaged across classes

| Feature | Accuracy | Balanced acc. | Precision | Recall | F1-score | Cohen's kappa |
|---|---|---|---|---|---|---|
| Hair color | 0.750 | 0.613 | 0.607 | 0.613 | 0.609 | 0.547 |
| Skin phototype | 0.727 | 0.730 | 0.566 | 0.584 | 0.566 | 0.626 |
| Hair thickness | 0.797 | 0.804 | 0.840 | 0.804 | 0.819 | 0.676 |
| Hair density | 0.781 | 0.726 | 0.842 | 0.726 | 0.740 | 0.634 |

hairs/mm$^2$). It is important to note that this discretization is applied only for this validation exercise; the system's final output remains numerical, allowing for flexible, application-specific thresholding.

The quantitative performance metrics across the test set are summarized in Table 7. For hair thickness and density, the model achieved balanced accuracies of 80.4% and 72.6%, with corresponding Cohen's Kappa scores of 0.676 and 0.634. Hair color classification yielded a balanced accuracy of 61.3% (Kappa = 0.547), while skin phototype classification reached 73.0% accuracy (Kappa = 0.626).

To analyze error distributions and patterns, confusion matrices for each feature are provided in Fig. 6. The matrices show that misclassifications occur predominantly between adjacent classes, while confusions between the extreme categories are zero. In the hair color matrix (Fig. 6a), the largest class-specific discrepancy occurred in the blond category, where 6 out of 8 ground-truth cases were predicted as brown.

Furthermore, to visualize the correspondence between the model's numerical predictions and the expert's categorical labels, boxplots were generated for hair thickness and density (see Fig. 7). These plots show a monotonic increase in the distribution of the model's continuous measurements across the expert's ordered classes, indicating that the numerical outputs scale consistently with the categories. Nevertheless, a partial overlap is visible in the tails of the boxplots, which corresponds to the adjacent-class confusions observed in the confusion matrices.

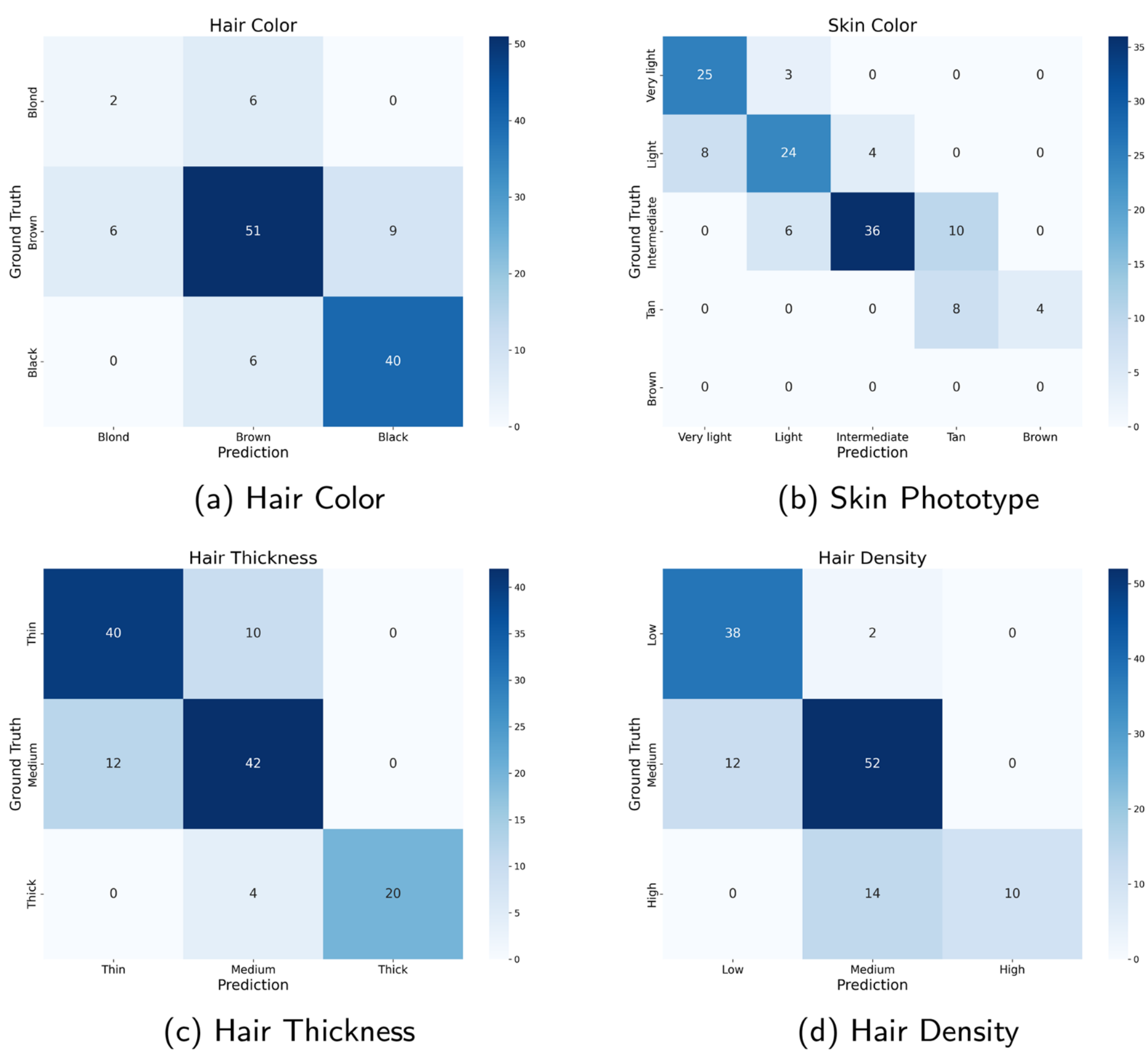

**Fig. 6** Confusion matrices for the classification of **a**) hair color, **b**) skin phototype, **c**) hair thickness, and **d**) hair density, comparing model predictions against expert annotations on the 128-image test set

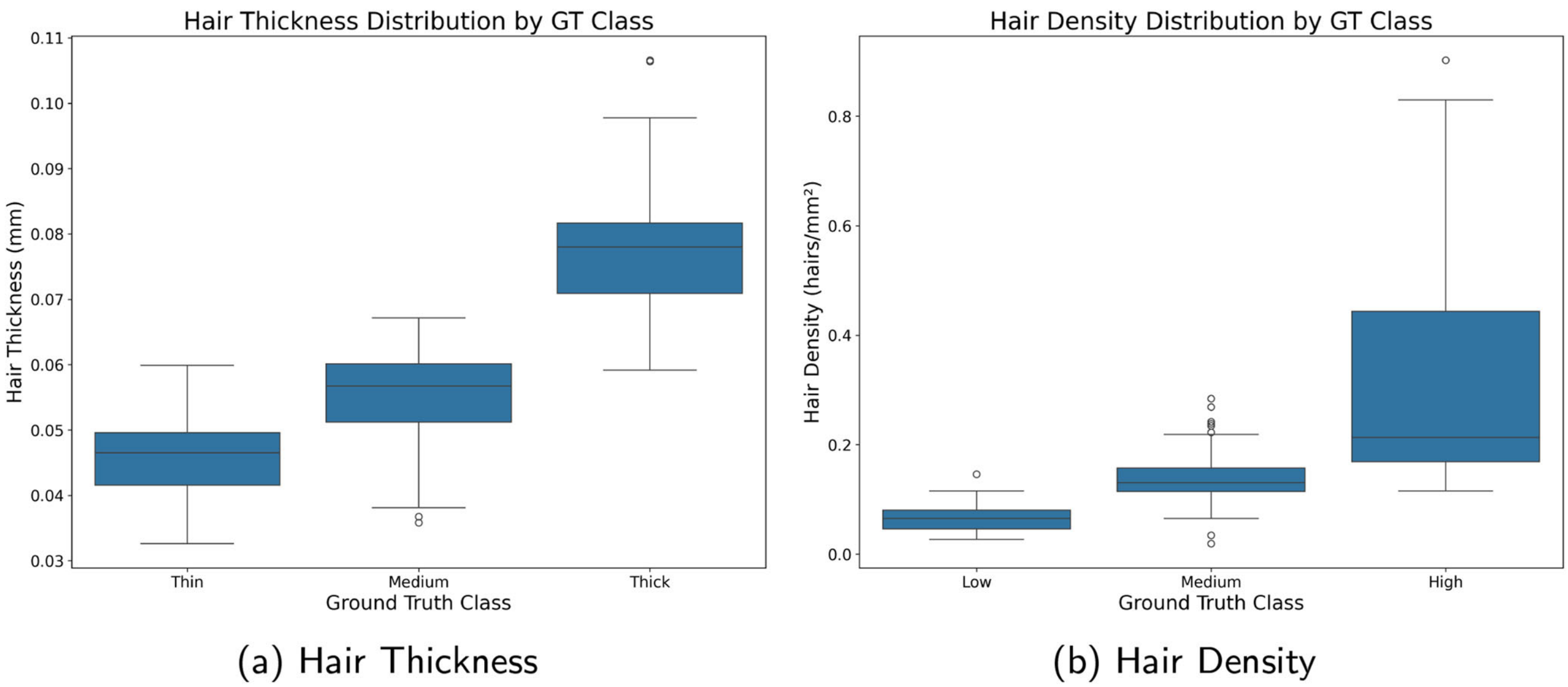

**Fig. 7** Boxplots showing the distribution of the model's continuous numerical predictions for **a**) hair thickness and **b**) hair density, grouped by the expert's ground truth categories (128-image test set). The plots demonstrate a clear monotonic trend, confirming the alignment between numerical predictions and categorical labels

**Table 8** Performance metrics for color-based feature extraction without color calibration, evaluated against expert annotations on the 128-image test set

| Feature | Accuracy | Balanced acc. | Precision | Recall | F1-score | Cohen's kappa |
|---|---|---|---|---|---|---|
| Hair color | 0.508 | 0.407 | 0.395 | 0.407 | 0.330 | 0.161 |
| Skin phototype | 0.016 | 0.018 | 0.200 | 0.014 | 0.027 | −0.096 |

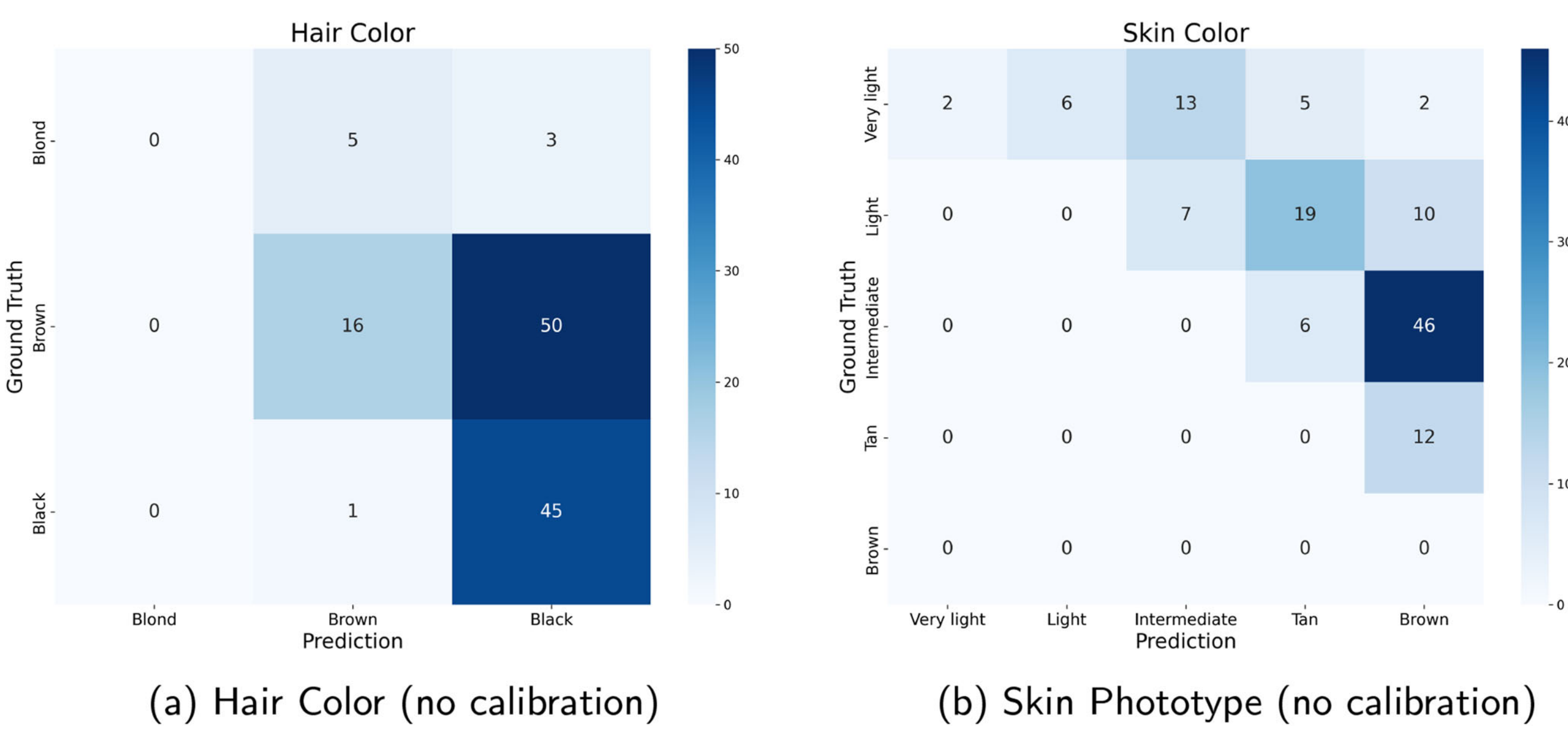

**Fig. 8** Confusion matrices for color-based features without calibration (on the 128-image test set), showing a collapse in predictive performance. **a**) Most hair is misclassified as black. **b**) Skin phototype predictions are systematically shifted towards darker classes

## Impact of Color Calibration

To demonstrate the critical role of the color calibration stage, an ablation study was conducted by disabling it and re-evaluating the color-dependent features. The results, presented in Table 8 and Fig. 8, reveal a catastrophic decline in performance.

Without calibration, skin phototype classification accuracy plummets from 72.7% to a mere 1.6%, with a negative Cohen's Kappa of −0.096, indicating that the model's performance is worse than random chance. The confusion matrix (Fig. 8b) confirms this failure, showing a systematic shift where predictions are displaced by approximately two classes towards darker phototypes, resulting in an almost empty main diagonal. Similarly, hair color accuracy drops from 75.0 to 50.8%, with the model incorrectly classifying the vast majority of hairs as black (Fig. 8a). As expected, the performance of geometric features like hair thickness and density remained unchanged, as they are not color-dependent.

## Full Pipeline Qualitative Results

To further illustrate the system's overall performance, Fig. 9 shows representative examples across different body areas and skin tones, providing a direct comparison between the model's continuous predictions and the expert's categorical ground truth (GT).

The segmentation stage accurately delineates even the shortest and least contrasted hair fragments, while the extracted attributes—including hair density, thickness, and color, as well as skin phototype—remain visually coherent and consistent across conditions. Despite illumination variations and subtle reflectance differences, the pipeline maintains stable results, supporting its potential for integration into real-world laser treatment systems.

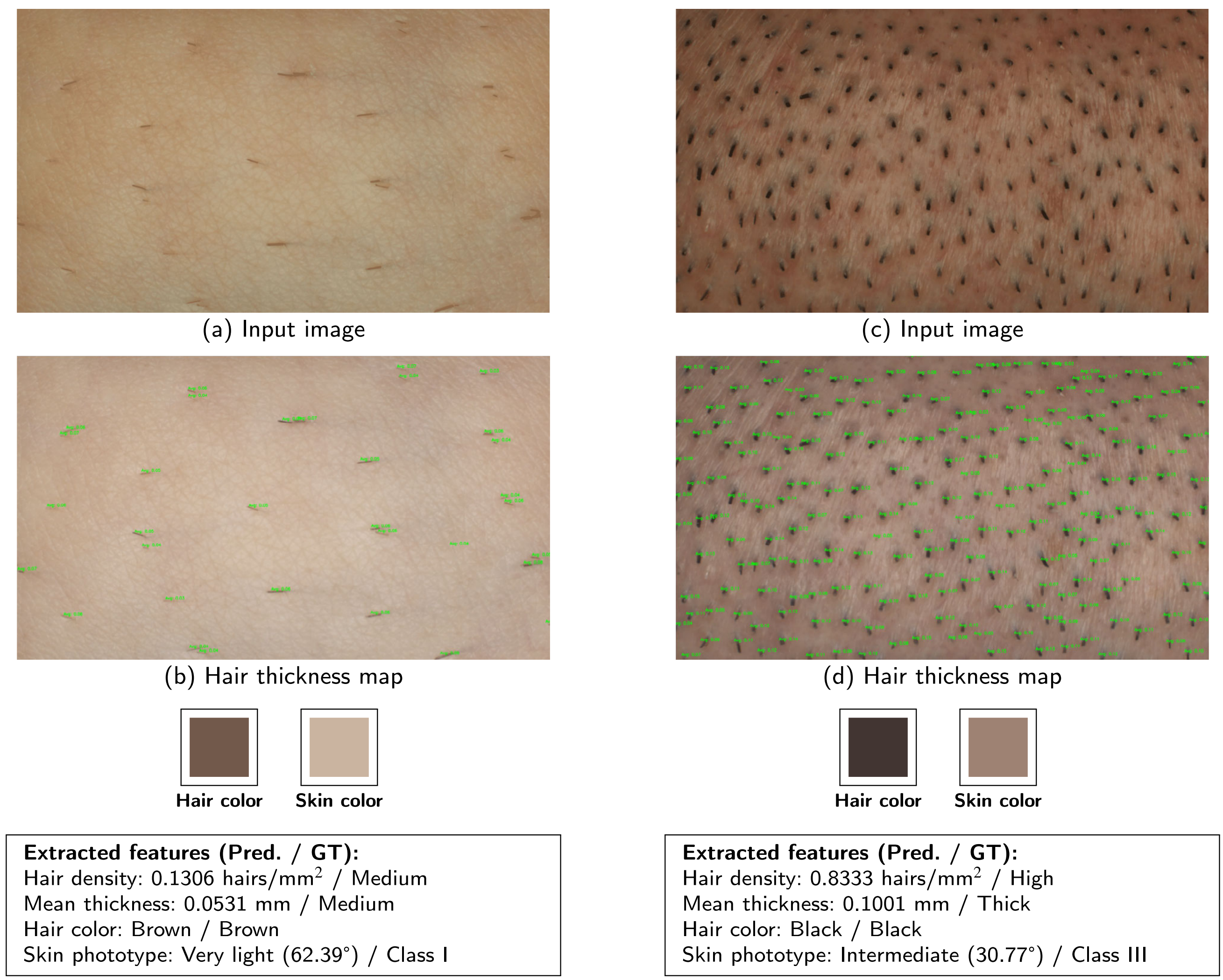

**Fig. 9** Full pipeline outputs for two representative samples. Columns show the input image (**a**, **c**) and thickness map (**b**,**d**), and extracted colors, comparing continuous predictions against the categorical ground truth (GT). Continuous metrics—density (number of hairs per $mm^2$) and thickness (in millimeters)—are computed over the full cropped area (2918 × 1664 pixels, ≈ 260.4 $mm^2$) using a calibrated spatial conversion factor of 0.007323 mm/pixel applied to the instance masks

## Discussion

The proposed pipeline demonstrates that fully automated characterization of the skin-hair interface is technically achievable. By combining the YOLOv9-C-DSEG instance segmentation model with a calibrated color processing pipeline, the system successfully bridges the gap between raw macroscopic imaging and objective clinical features, such as skin phototype, hair density, and thickness. It is important to emphasize that this work focuses strictly on the automated extraction of these characteristics as a preliminary diagnostic phase; the direct estimation of specific irradiation parameters falls outside the scope of this study and remains a subject for future clinical validation. Nevertheless, because these extracted features represent the primary biomarkers required by clinicians to calculate safe dosimetry, this robust characterization provides the mandatory prerequisite foundation for any subsequent automated parameterization workflow.

While these features are strictly necessary for determining optimal parameters in laser hair removal, the potential of this system extends significantly further. Current light-based devices, such as intense pulsed light (IPL), vascular laser therapy, and dermatologic phototherapy, are widely used in clinical dermatology and could benefit from this objective characterization. Extending the proposed approach to these medical areas where pulsed light is used would provide a more precise treatment resolution.

Furthermore, while the comparative architectural evaluation demonstrates that all candidates achieve similarly high localization performance ($mAP_{50} \approx 0.80$), the detailed mask analysis justifies the selection of YOLOv9-C-DSEG due to its superior boundary and region-based fidelity. The high micro-averaged scores confirm that the model remains highly resilient across the vast majority of hair instances globally, even though macro-averaged metrics reflect a higher sensitivity to isolated, low-contrast edge cases. The entire pipeline operates efficiently, reaching sub-second inference times on mid-range GPUs, which strongly supports its future integration into real clinical devices.

Regarding the quantitative feature evaluation, the subpixel absolute error (4.0 μm) in mean hair thickness reported in the "Error Propagation to Feature Estimation" section confirms that minor segmentation boundary variations do not critically propagate downstream, proving that the feature extraction pipeline is mathematically robust against mask noise. In contrast, the overlap in the tails of the boxplots and the confusion between adjacent classes in the matrices do not solely represent system error. Instead, they highlight the inherent subjectivity and variability of human categorical annotation, which is a core motivation for this work. The extremely low error reported when comparing against precise ground-truth masks supports the conclusion that this interclass confusion is largely attributable to annotator variability rather than calculation inaccuracies. By the same logic, it is probable that some confusion between adjacent color classes (e.g., brown vs. black) also stems from annotation noise. A notable exception is the misclassification of blond hair, which represents a limitation of the system. In some cases, hair fineness introduces an optical bias driven by the underlying skin tone. While fine dark hairs can appear lighter due to skin 'show-through', highly translucent fine blond hairs suffer from the opposite effect: the underlying skin color and microshadows darken their perceived chromatic profile, causing them to be incorrectly classified as brown. Finally, the catastrophic drop in performance observed when removing the color correction stage (where skin phototype accuracy collapsed to a mere 1.6%) underscores that spectrophotometric calibration is not a minor enhancement, but a strict operational requirement to neutralize camera-induced biases and ambient lighting fluctuations.

Nonetheless, certain limitations should be acknowledged. Although the annotated dataset covers all defined classes, its limited size—inherently constrained by the number of available volunteers—may not fully represent the precise data distribution of the general population. While expanding the dataset would be beneficial to further consolidate the system, the current data provides a highly reliable baseline for this methodology. A specific challenge arises with very dark skin tones, where the inherent lack of contrast between hair and skin prevents reliable detection using standard macroscopic imaging. Finally, it should be noted that the proposed color calibration stage is specifically designed for a controlled acquisition setup; therefore, it cannot be directly transferred to arbitrary or uncalibrated capture devices without prior adaptation.

## Conclusion and Future Work

This study validates the feasibility of combining deep learning-based segmentation and calibrated image analysis for the automatic, objective characterization of macroscopic dermatological images. The integrated system provides reproducible quantitative estimations of key skin and hair features, offering a robust, objective alternative to subjective manual feature assessment prior to laser treatments. Crucially, the experimental evaluation confirms both the sub-pixel resilience of the downstream feature extraction against segmentation noise and the strict necessity of spectrophotometric calibration to ensure viability under lighting biases and fluctuations.

While the current results are highly promising, extensive validation through larger datasets and controlled studies is necessary to confirm its clinical applicability. Future work will focus on scaling data collection—ensuring a representative distribution of the clinical population that encompasses all defined classes and incorporating multi-expert consensus protocols to mitigate annotation subjectivity and noise—and conducting clinical trials to evaluate the real-world impact of automated characterization on treatment safety and personalization. To address current technical limitations, future efforts will also explore alternative instance segmentation architectures, such as Mask R-CNN or attention-based transformers, to determine if boundary delineations for extremely fine structures can be further optimized, as well as thickness-conditioned hair color classification frameworks to resolve the misclassification of very light or fine hairs caused by transparency and focus blur. Additionally, alternative imaging modalities will be evaluated to improve detection in dark skin tones, alongside the implementation of self-calibrating illumination or improved optical uniformity. Ultimately, the presented approach establishes a strong foundation for the development of next-generation intelligent laser systems capable of adaptive, patient-specific parameterization.

**Acknowledgements** The authors would like to thank Intek Medical Technology and the Universitat Politècnica de Catalunya (UPC) for providing the environment and resources that made this research possible.

**Author Contributions** Alex Navarro Navarro: conceptualization, methodology, software, writing—original draft. Marc Arriaga Hidalgo: investigation, data curation, writing—review and editing. Carlos Moreno Rubio: investigation, data curation, writing—review and editing. Maxime Talarmain: investigation, data curation, writing—review and editing. Noelia Vaquero Gallardo: conceptualization, investigation, writing—review and editing, supervision.

**Funding** Open Access funding provided thanks to the CRUE-CSIC agreement with Springer Nature. This work was supported by the Fondo Europeo de Desarrollo Regional (FEDER) through the Centro para el Desarrollo Tecnológico y la Innovación (CDTI) [grant number IDI-20240216].

**Data Availability** Data are not publicly available due to commercial confidentiality and volunteer privacy.

## Declarations

**Ethics Approval** This study was conducted in accordance with the ethical standards of the 1964 Declaration of Helsinki and its later amendments or comparable ethical standards. The study involved only the non-invasive acquisition of non-identifiable photographs of hair from adult volunteer participants for the development and evaluation of computer vision models aimed at optimizing laser hair removal parameters. No clinical procedures, medical interventions, or collection of personally identifiable information were performed. Formal ethics review and approval by an independent ethics review committee or equivalent body was not required for this study; therefore, no ethics approval number was applicable.

**Consent to Participate** Informed consent was obtained from all individual participants included in the study. Written informed consent was obtained from all participants prior to image acquisition and use of the photographs for research purposes, in accordance with applicable data protection requirements, including the General Data Protection Regulation (EU) 2016/679 and Spanish Organic Law 3/2018. The consent form informed participants about the purpose and procedures of the study, the use and protection of their images, their right to withdraw consent, and the applicable data protection measures. All images were fully anonymized prior to analysis and were intended exclusively for research purposes.

**Consent for Publication** All authors have approved the manuscript and agree with its submission to the Journal of Imaging Informatics in Medicine. Furthermore, the authors affirm that human research participants provided informed consent for publication of the macroscopic images included in the manuscript.

**Competing Interests** The authors declare no competing interests.